\documentclass[conference]{IEEEtran}
\IEEEoverridecommandlockouts
\usepackage{cite}
\usepackage{amsmath,amssymb,amsfonts}
\usepackage{algorithmic}
\usepackage{graphicx}
\usepackage{textcomp}
\usepackage{xcolor}

\usepackage{graphicx}
\usepackage{float}
\usepackage{subfig}
\usepackage{multirow}
\usepackage{multicol}
\usepackage{amssymb}
\usepackage{xcolor}
\usepackage{amsmath}
\usepackage{bbding}
\usepackage{colortbl}
\definecolor{lightgray}{rgb}{0.86, 0.86, 0.86}

\def\BibTeX{{\rm B\kern-.05em{\sc i\kern-.025em b}\kern-.08em
    T\kern-.1667em\lower.7ex\hbox{E}\kern-.125emX}}
\begin{document}

\title{ActionPrompt: Action-Guided 3D Human Pose Estimation With Text and Pose Prompting
}

\author{\IEEEauthorblockN{Hongwei Zheng\IEEEauthorrefmark{1}, Han Li\IEEEauthorrefmark{1}, Bowen Shi\IEEEauthorrefmark{1}, Wenrui Dai\IEEEauthorrefmark{1}\thanks{Correspondence to Wenrui Dai.}, Botao Wang\IEEEauthorrefmark{2}, Yu Sun\IEEEauthorrefmark{2}, Min Guo\IEEEauthorrefmark{2}, Hongkai Xiong\IEEEauthorrefmark{1}
}

\IEEEauthorblockA{\IEEEauthorrefmark{1}Shanghai Jiao Tong University, Shanghai, China\\\
Email: \{1424977324, qingshi9974, sjtu\_shibowen, daiwenrui, xionghongkai\}@sjtu.edu.cn}
\IEEEauthorblockA{\IEEEauthorrefmark{2}Qualcomm AI Research\thanks{Qualcomm AI Research is an initiative of Qualcomm Technologies, Inc. Datasets were downloaded and evaluated by Shanghai Jiao Tong University researchers.}, Shanghai, China\\
Email: \{botaow, sunyu, mguo\}@qti.qualcomm.com}}

\maketitle

\begin{abstract}
Recent 2D-to-3D human pose estimation (HPE) utilizes temporal consistency across sequences to alleviate the depth ambiguity problem but ignore the action related prior knowledge hidden in the pose sequence. In this paper, we propose a plug-and-play module named \textbf{A}ction \textbf{P}rompt \textbf{M}odule (APM) that 
effectively mines different kinds of action clues for 3D HPE. The highlight is that, the mining scheme of APM can be widely adapted to different frameworks and bring consistent benefits.
Specifically, we first present a novel Action-related Text Prompt module (ATP) that directly embeds action labels and transfers the rich language information in the label to the pose sequence.
Besides, 
we further introduce Action-specific Pose Prompt module (APP) to mine the position-aware pose pattern of each action, and exploit the correlation between the mined patterns and input pose sequence for further pose refinement.
Experiments show that APM can improve the performance of most video-based 2D-to-3D HPE frameworks by a large margin. 

\end{abstract}


\section{Introduction}
\begin{figure}[!t]
\centering
\subfloat[\label{fig:motivation2}]{\includegraphics[scale=0.28]{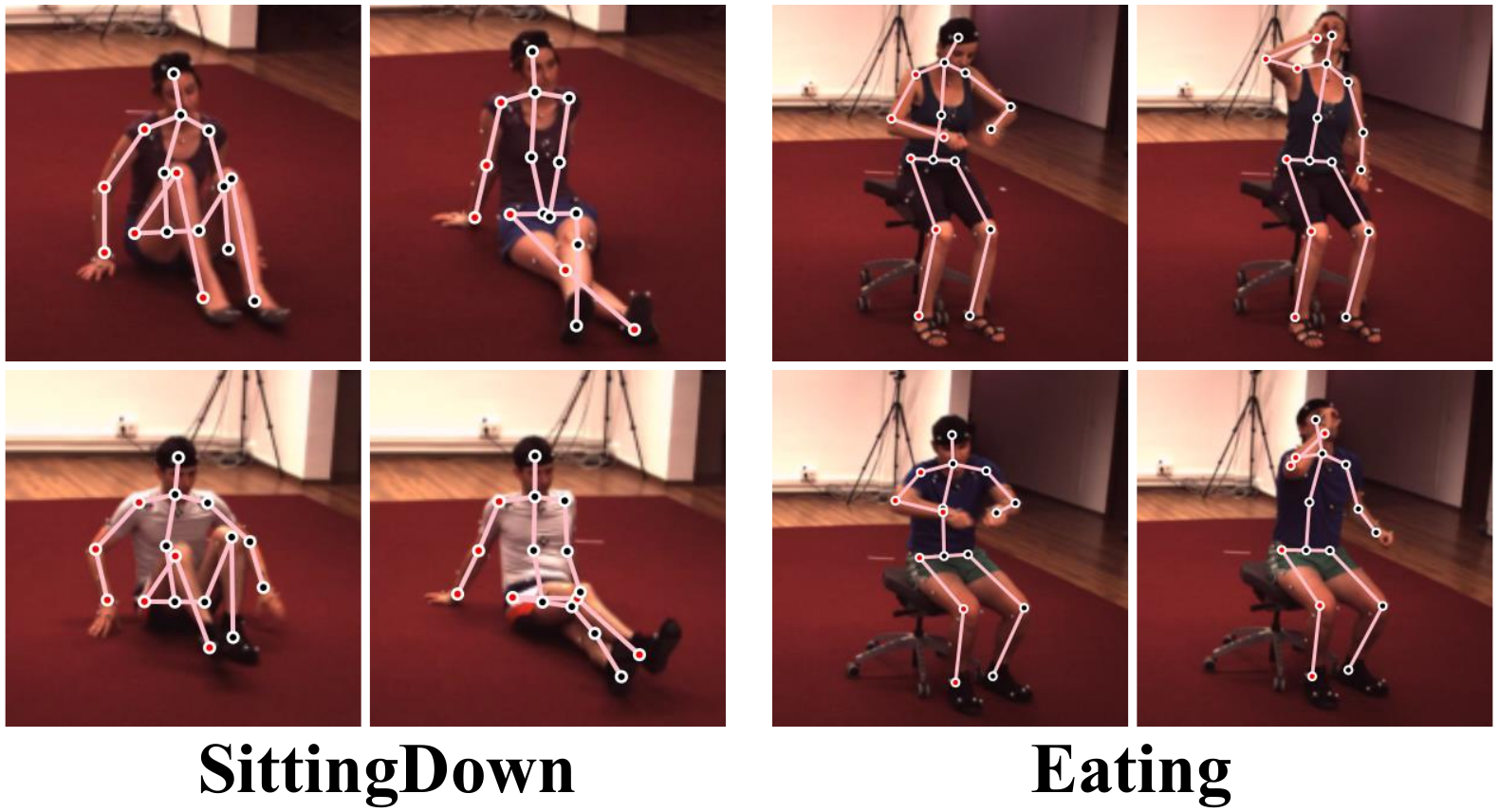}}\vspace{-10pt}\\
\subfloat[\label{fig:motivation1}]{\includegraphics[scale=0.46]{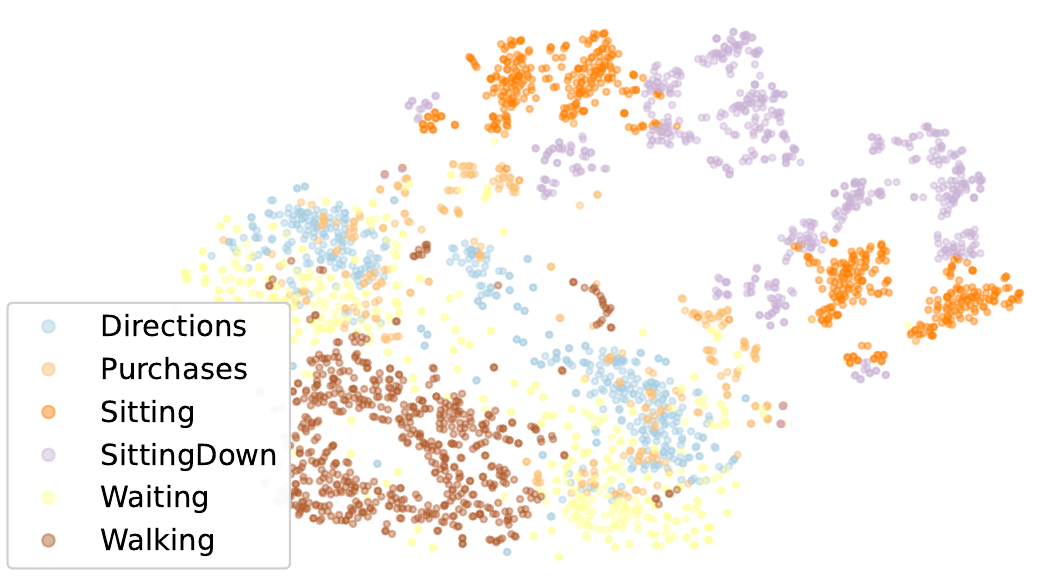}}
\caption{(a) Examples in Human3.6M. (b) 3D pose distribution of several actions in Human3.6M dataset by t-SNE~\cite{van2008visualizing}.}\label{fig:motivation}
\vspace{-10pt}
\end{figure}

3D human pose estimation (HPE) from a monocular image or video has been widely considered in a variety of applications in human action recognition, robotics, and human-computer interaction. 3D HPE usually follows a 2D-to-3D pipeline that first estimates 2D joints from the input image and then lifts the 2D joints to 3D pose. However, due to the absence of depth information, this pipeline suffers from the serious depth ambiguity problem~\cite{zeng2021learning,li2021hierarchical,li2023pose} caused by the many-to-one mapping from multiple 3D poses to one same 2D projection.  

Recent attempts~\cite{pavllo20193d,liu2020attention,zhang2022mixste,zheng20213d,zeng2022smoothnet} exploit the temporal consistency across sequences to alleviate these problems. 
However, they only model the action-agnostic spatial and temporal correlations but ignore the action related prior knowledge contained in the pose sequence. As shown in Fig.~\ref{fig:motivation}(a), depth ambiguity is more likely to occur in the part of \emph{feet} for the action \emph{SittingDown}, whereas in the part of \emph{hands} for the action \emph{Eating}, since these parts usually have large motion. Furthermore, Fig.~\ref{fig:motivation}(b) shows that 3D pose distributions significantly differ for different actions in Human3.6M~\cite{ionescu2013human3}. This fact suggests that each action has unique characteristics that could benefit pose estimation. It is necessary to effectively embed the action clues into video-based 2D-to-3D pose estimation.

Existing methods form a multi-task learning framework to embed the action clues by simultaneously considering pose estimation and action recognition.
In ~\cite{luvizon20182d,luvizon2020multi,liu2020learning}, the action projector is introduced to classify the pose sequence with one-hot action label but yields trivial performance gain in pose estimation. 
We argue that this is because the action clues cannot be fully mined by simply using the one-hot action label in the classification sub-task.
One-hot action label only contains overall movement category information but ignores the position and velocity information of the pose sequence. \textbf{Therefore, a more effective mining scheme for action label needs to be carefully designed rather than such one-hot manner.}

In this paper, we propose a plug-and-play module named \textbf{A}ction \textbf{P}rompt \textbf{M}odule (APM) that can mine different kinds of action clues into HPE for better feature extraction. Motivated by recent vision-language models (VLM) like CLIP~\cite{radford2021learning}, which utilizes a huge amount of image-text pairs for pretraining and can benefit the visual features with additional information from text, we first propose a novel Action-related Text Prompt module (ATP) that embeds each action label into text prompts for the pose sequence feature enhancement.
Considering that CLIP pretrained model may lack knowledge about pose sequence, we also design a Pose-to-text Prompt (P2T) module in ATP to endow the text prompt embeddings with velocity information. 
After obtaining the action-related text prompts, we align the feature of pose sequence with its corresponding prompt. In this way, the rich action-related language information can be transferred to the pose sequence. 
It's worth noting that our ATP makes the first attempt to leverage action-related language information from pre-trained VLM model for HPE.

Though promising it is, text prompts of actions still lack some position-aware information about human pose.
Inspired by our finding that some representative poses for the same action are shared across different subjects, as shown in Fig~\ref{fig:motivation}(a), we further propose an Action-specific Pose Prompt module (APP) to effectively mine and exploit these action-specific pose patterns, which are position-aware and shared across different subjects.  
In particular, for each action, we utilize some pose prompts as a learnable action-specific pattern to capture typical position-aware information. Then we perform cross attention by regarding feature of pose sequence as query, and these pose prompts as key and value, thus matching the learnable action-specific representative position-aware information with input pose sequence and refine pose feature to obtain more accurate 3D estimated pose.

The proposed APM is a general plug-and-play module to improve existing video-based 2D-to-3D pose estimation models. 
We seamlessly employ our method to three recent state-of-the-art models, including VPose~\cite{pavllo20193d}, A3DHP~\cite{liu2020attention} which are the classic model using temporal convolutions~\cite{lea2017temporal} and MixSTE ~\cite{zhang2022mixste} which is the current SOTA model based on Transformer~\cite{vaswani2017attention}.
The proposed APM is shown to improve all the models on the Human3.6M and HumanEva-I datasets. Remarkably, it achieves an average gain of more than 5\% in MPJPE for all the models. Furthermore, the proposed APM alleviates the depth ambiguity of different actions, especially for hard actions. 

\begin{figure*}
\renewcommand{\baselinestretch}{1.0}
\centering
\includegraphics[scale=0.6]{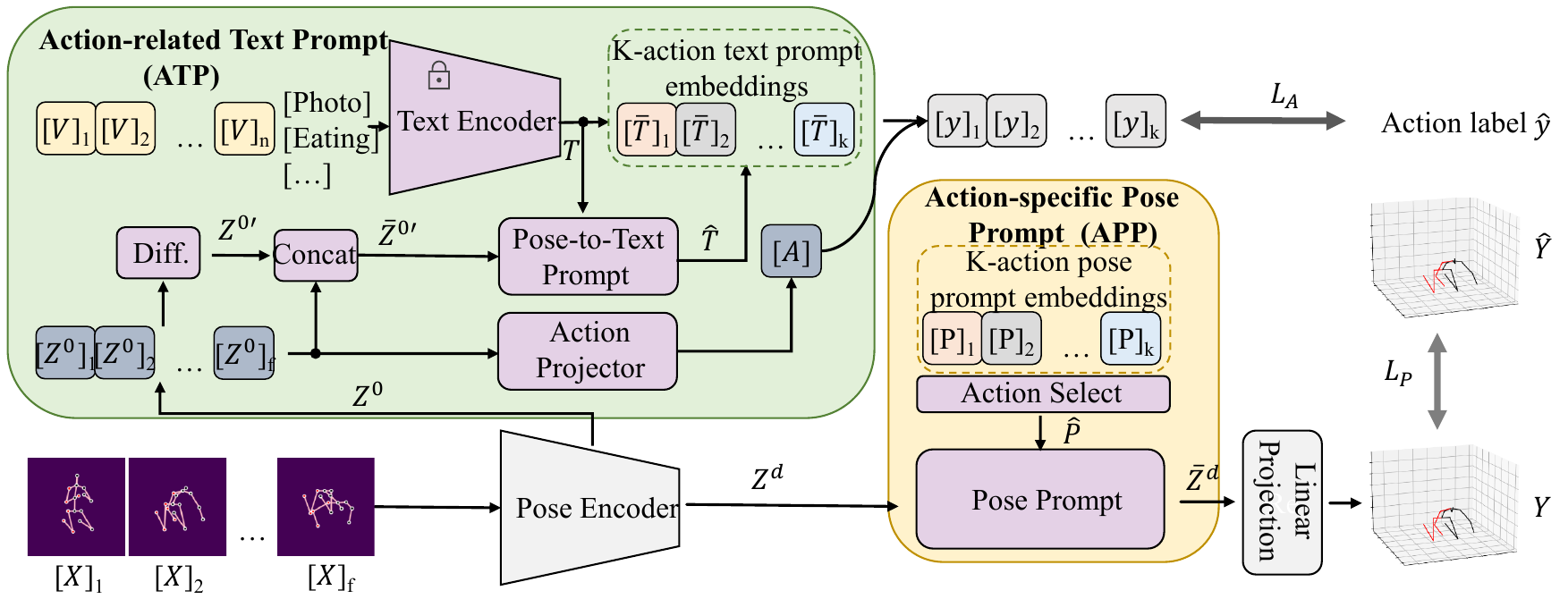}
\caption{The overview of Action Prompt Module (APM), which is a plug-and-play module consisting of two sub-modules, i.e.,  Action-related Text Prompt (ATP) and Action-specific Pose Prompt (APP). Taken 2D pose sequence as the input of pose encoder, ATP directly embeds action labels and transfers the rich action-related language information to the pose sequence by aligning poses with texts in the feature space. Then, APP mines the position-aware pose pattern of each action   and exploits the correlation between the patterns and pose feature for further refinement.
Finally, we predict the target 3D pose with refined pose feature. The architecture of pose encoder and linear projection is according to baseline model. (The text encoder is frozen during training and is abandoned during inference.)}\label{fig_network_2}
\end{figure*}

\section{Action Prompt Module}
Fig.~\ref{fig_network_2} illustrates the proposed plug-and-play Action Prompt Module for video-based 2D-to-3D HPE models. APM consists of two sub-modules, i.e., Action-related Text Prompt (ATP) and Action-specific Pose Prompt (APP).
Taken 2D pose sequence as the input of pose
encoder, ATP directly embeds action labels and transfers the action-related language information to the pose sequence by aligning poses with texts in the feature space. APP mines the position-aware pose pattern of each action and exploits the correlation between the patterns and pose feature for further refinement. The refined pose feature is leveraged to predict the target 3D pose.

\subsection{Action-related Text Prompt}
ATP exploits the rich action-related language information to align the pose features with text features. 
Given the input 2D pose sequence $X\in \mathbb{R}^{F\times J\times 2}$ that contains $F$ frames and $J$ joints for each frame, we obtain the $C$-dimensional pose features of the shallow layer $Z^0\in \mathbb{R}^{F\times C}$ in the pose encoder and then project them into the action features $A \in \mathbb{R}^{C}$ by Action Projector. 
Meanwhile, the learnable text prompts $V \in \mathbb{R}^{K\times N\times C}$, where $K$ denotes the number of action classes and $N$ denotes the number of text prompts, are encoded by the text encoder to obtain $K$-action text prompt embeddings $T  \in \mathbb{R}^{K\times C}$ containing action-related information. In addition, Pose-to-Text Prompt endows the text prompt embedding $T$ with velocity information contained in the pose features $Z^0$. Finally, we can get the classification vector $y\in\mathbb{R}^{K}$ calculated between the enhanced text prompt embedding $\bar{T}$ and action feature $A$ for feature alignment. 
As highlighted in green in Fig.~\ref{fig_network_2}, ATP includes the text prompt design for 3D HPE, Action Projector, and Pose-to-Text Prompt. 

\noindent\textbf{Design of Text Prompt.} We firstly introduce text prompt design to embed action label in a learnable manner.
Inspired by CoOp~\cite{zhou2022learning}, we  utilize  learnable text prompts as our templates by optimizing them in the training process. In order to adapt to 3D HPE, we fill up the [CLS] token with various action class names in text prompts. The input of the text encoder can be formulated as:
\begin{equation}
V = [V]_1[V]_2...[V]_N[Action]_k,\quad k=1,...,K,
\end{equation}
where $[V]_i\in\mathbb{R}^C$ is the randomly initialized learnable template and is shared by all action classes and $[Action]_{k}\in\mathbb{R}^C$ represents the corresponding action class. 
The text prompts $V$ are input to the text encoder, whose parameter is frozen except for the final text projection layer during training. Then, we can obtain the $K$-action text prompt embeddings $T$, containing rich action-related language information.

\noindent\textbf{Action Projector.} We propose the Action Projector to extract the action-related information contained in the pose features. 
As pose features propagate through the VPose encoder, the time dimension of pose features reduces from $F$ to $1$,
which means that the temporal information is lost as the network deepens. 
Therefore, 
we connect the Action Projector with the the shallowest layer of the pose encoder that contains more general information for action recognition, and get action features $A$, i.e., $A=\texttt{Proj}(Z^0)$. The Action Projector is realized using $D_2$ TCN~\cite{lea2017temporal} blocks.

\noindent\textbf{Pose-to-Text Prompt.} Inspired by DenseCLIP~\cite{rao2022denseclip}, we propose the Pose-to-Text Prompt module, which endows the text prompt embeddings with velocity information via cross attention layer, thus making up the lack of knowledge about pose sequence in pretrained CLIP model. In that case, the enhanced text prompts can describe the action more accurately and concretely. For example, the text prompt ``a video of walking fast'' is more accurate than ``a video of walking", as it incorporates the velocity information of pose movement. 
Thus, we extract the first-order motion information of pose features $Z^{0\prime}\in\mathbb{R}^{(F-1) \times C}$, which is the difference of neighbor frames of $Z^0$, to represent the velocity and concatenate the pose features $Z^0$ and $Z^{0\prime}$ to obtain $\bar{Z^0}$, i.e., $\bar{Z}^0=\texttt{Concat}(Z^0,Z^{0\prime})$. Subsequently, the text prompt embedding $T$ is fed into the cross-attention as queries while $\bar{Z}^0$ as keys and values to obtain the output $\hat{T}$. In that way, the text features can find the most related pose clues. The enhanced text prompt embedding $\bar{T}$ is achieved by combining $\hat{T}$ and $T$.
\subsection{Action-specific Pose Prompt}
We develop the Action-specific Post Prompt (APP) to address the problem that text prompts with action label still lack position-aware information about human pose. APP mines the position-aware pose pattern of each action, and exploits the correlation between the mined patterns and input pose sequence for pose refinement.


In APP, we propose the learnable pose prompts $P\in\mathbb{R}^{K\times L\times C}$ as learnable action-specific patterns, where $L$ is the number of pose prompts for each action. It is worth noting that there are $K$-class pose prompt templates which are action-specific because they are designed to learn more fine-grained information compared with the text prompts. 
During training, the pose prompts $\hat{P}\in\mathbb{R}^{L\times C}$ are selected according to the given action label. Then we perform cross attention by regarding the output pose features of the pose encoder $Z^{d}\in \mathbb{R}^{1\times C}$ as query, and select pose prompts $\hat{P}$ as key and value in the Transformer decoder~\cite{vaswani2017attention}.
\begin{equation}
\hat{Z}^{d}=\texttt{TransDecoder}(Z^{d},\hat{P}).
\end{equation}
In this case, the pose features are matched with the most related pose pattern of the corresponding action. Similarly, we refine the pose features through the residual connection:
\begin{equation}
\bar{Z}^{d}=Z^{d}+\gamma\hat{Z}^{d},
\end{equation}
where $\gamma\in\mathbb{R}^C$ is the learnable parameter to scale the residual $\hat{Z}^{d}$. Finally, the target 3D pose $Y \in \mathbb{R}^{J \times 3}$ is obtained from the refined pose feature $\bar{Z}^{d}$ through linear projection. 

Note that the text encoder of ATP is only needed in the training process, which facilitates model deployment in the real world. During inference, the optimized text prompt embeddings of all the actions are saved locally to infer action label, thus APP can select the pose prompts of corresponding action to refine pose features. 

\begin{table*}[!t]
\renewcommand{\baselinestretch}{1.0}
\renewcommand{\arraystretch}{1.0}
\setlength\tabcolsep{3pt}
\begin{center}
\caption{Quantitative evaluation results of Action Prompt Module attached to various pose encoders on Human3.6M (with an input length $F$ of 243). The inputs for the top group are ground truth (GT) of 2D pose while the bottum group are the detection 2D pose from HRNet~\cite{sun2019deep}(denoted by $^\ast$). The per-action results are shown in P1 and avgerage P1, P2, P3 are calculated.}
\label{tab:res}
\begin{tabular}{@{}l|ccccccccccccccc|ccc@{}}
\hline
Method&Walk &	WalkT. &	Eat &	Pur. &	WalkD. &	Phone &	Smoke &	Greet &	Dir. &	Wait &	Photo &	Disc. &	Pose &	SitD. &	Sit &P1 & P2 & P3\\
\hline
\hline
VPose &29.3&	30.0	&34.8	&37.8&	37.0&	35.5&	36.9&	36.3&	36.9&	37.7&	40.2&	41.1&	41.4&	45.9&	45.8& \multicolumn{1}{l}{37.8}& \multicolumn{1}{l}{33.6}& \multicolumn{1}{l}{39.3} \\
+APM &26.6 &	27.1 &	34.0 &	35.8 &	36.1 &	35.1 &	34.8 &	36.0& 	34.9 &	33.9 &	39.1 &	37.2 &	39.2 &	43.4 &	42.0 &	35.7\textcolor{red}{(-2.1)}& 29.7\textcolor{red}{(-3.9)}&33.8\textcolor{red}{(-5.5)}\\
\hline
A3DHP&26.2 &	26.8&	33.5 &	34.5 &	35.7 &	37.4 &	35.5 &	35.3 &	34.3 &	39.1 &	44.9 &	39.1 &	39.7 &	49.5 &	46.4&	\multicolumn{1}{l}{37.2}&  \multicolumn{1}{l}{33.4}& \multicolumn{1}{l}{41.2}\\
+APM& 24.8 &	26.4 &	31.4 &	31.7 &	35.1 &	36.4 &	35.0 &	33.6 &	31.5 &	36.4 &	44.5 &	37.1 &	37.4 &	49.5& 	43.8 & 35.6\textcolor{red}{(-1.6)} &31.9\textcolor{red}{(-1.5)} &38.8\textcolor{red}{(-2.4)}\\
\hline
MixSTE &15.7 &	16.1 &	22.4 &	23.3 &	23.2 &	23.1 &	23.5 &	23.1 	&23.6 &	24.1 &	28.5 &	24.3& 	26.5 &	32.0 &	30.2 &	\multicolumn{1}{l}{24.0} & \multicolumn{1}{l}{21.8} & \multicolumn{1}{l}{27.5}\\
+APM& 15.9 &	16.7&	21.1 	&22.5 &	22.1 &	21.8 &	22.7 &	21.9 &	22.3 &	23.0 &	25.8 &	21.8 &	25.1 &	30.2 &	27.9 & 22.7\textcolor{red}{(-1.3)} & 19.9\textcolor{red}{(-1.9)} & 24.7\textcolor{red}{(-2.8)}\\
\hline
\hline
VPose$^\ast$ &37.0&	38.8&	43.4&	42.6&	49.1&	49.3&	45.9	&47.4&	44.5&	45.2&	53.4&	47.0&	46.3&	67.6&	55.6	& \multicolumn{1}{l}{47.5}&	\multicolumn{1}{l}{34.6}&	\multicolumn{1}{l}{41.7}\\
+APM$^\ast$ &36.7&	38.7&	43.5&	41.7&	49.3&	48.7&	45.0&	47.2&	42.2&	43.8&	52.1&	45.2&	44.3&	65.3&	53.5	&46.5\textcolor{red}{(-1.0)}&	33.3\textcolor{red}{(-1.3)}&	39.8\textcolor{red}{(-1.9)}\\
\hline
A3DHP$^\ast$ &36.1	&38.3&	43.5&	41.9&	47.9&	50.0&	45.7&	47.2&	43.4&	46.1&	56.9&	45.6&	45.6&	71.0&	56.9&	\multicolumn{1}{l}{47.7}&	\multicolumn{1}{l}{35.3}&	\multicolumn{1}{l}{43.7}\\
+APM$^\ast$ &35.8&	38.2&	43.5&	41.1&	47.4&	49.9&	45.7&	46.2&	42.8&	45.7&	56.1&	44.7&	43.7&	68.5&	54.5	&46.9\textcolor{red}{(-0.8)}&	34.2\textcolor{red}{(-1.1)}&	42.0\textcolor{red}{(-1.7)}\\
\hline 
MixSTE$^\ast$ &31.8	&32.8&	40.6&	38.3&	42.6&	43.8&	42.0&	40.9&	37.6&	40.5&	49.1&	40.3&	39.4&	65.2&	52.9&	\multicolumn{1}{l}{42.5}&	\multicolumn{1}{l}{30.9}&	\multicolumn{1}{l}{38.9}\\
+APM$^\ast$ &30.5&	31.6&	39.8&	38.9&	42.0&	43.1&	42.0&40.6&	37.7&	40.0&	48.0&	40.2&	38.8&	63.2&	50.8&	41.8\textcolor{red}{(-0.7)}&	29.5\textcolor{red}{(-1.4)}&	36.8\textcolor{red}{(-2.1)}\\
  \hline
\end{tabular}
\end{center}
\end{table*}
\subsection{Training Loss}
The overall training loss $\mathcal{L}=L_P+\lambda\cdot L_A$ balances the pose loss $L_P$ and action loss $L_A$ with a trade-off factor $\lambda$. 

\noindent \textbf{Pose Loss.} The pose loss $L_P$ is formulated as
\begin{equation}
L_P = \frac{1}{J}\sum_{i=1}^{J}\Vert {\hat{Y}_i - Y_i} \Vert_2,
\end{equation}
where $\hat{Y}_i$  and $Y_i$ are respectively the ground truth and estimated 3D joint location of the $i$-th joint.

\noindent\textbf{Action Loss.} The classification vector $y\in\mathbb{R}^{C}$ is predicted using the cosine similarity between the text prompt embedding $\bar{T}$ and the action feature $A$.
\begin{equation}
p(y=i|x) = \frac{\exp(\cos(\bar{T}_i,A )/\tau)}{\sum_{j=1}^{K}\exp(\cos(\bar{T}_j,A )/\tau)},
\end{equation}
where $\tau$ is a temperature parameter.
We then define the action loss $L_A$ as the cross-entropy loss between the ground truth $\hat{y}$ and predicted classification vector $y$.
\begin{equation}
L_A = \texttt{CrossEntropy}(\hat{y}, y).
\end{equation}


\section{Experiments}
\subsection{Experimental Settings}
\textbf{Datasets.} Human3.6M~\cite{ionescu2013human3} is the most commonly used indoor dataset for 3D HPE that contains 3.6 million images of 11 subjects and 15 actions. Following~\cite{pavllo20193d,liu2020attention, zhang2022mixste}, we take five subjects (S1, S5, S6, S7, S8) for training and another two subjects (S9, S11) for testing. We evaluate our method and conduct ablation studies on the Human3.6M. HumanEva-I~\cite{sigal2010humaneva} is further adopted to demonstrate the generalization ability of the proposed method. It consists of seven calibrated sequences for four subjects performing six actions. Following~\cite{liu2020attention, zhang2022mixste}, we test our models on the actions of Walk and Jog.

\noindent\textbf{Evaluation Metrics.}
Follow previous work~\cite{pavllo20193d,liu2020attention, zhang2022mixste}, we use the mean per-joint position error (MPJPE) as evaluation metric. To evaluate the effect on the alleviation of depth ambiguity, we also provide the position error of depth axis, termed as D-MPJPE.
In addition, we calculate the D-MPJPE of three hardest actions (Posing, SittingDown and Sitting), termed as Tail D-MPJPE, to focus on these actions with significant error. In the following parts, We abbreviate MPJPE, D-MPJPE, Tail D-MPJPE as P1, P2 and P3, respectively.

\noindent\textbf{Implementation Details.} The proposed method is implemented with PyTorch. The text encoder in APP loads weight from the pretrained text encoder in CLIP~\cite{radford2021learning}. To demonstrate the effect of the proposed method, we apply APM to several existing video-based 2D-to-3D HPE methods, including VPose, A3DHP, and MixSTE. $\lambda$ is set to 0.1. For fair comparison, we apply the same parameter settings as the corresponding baseline experiments in~\cite{pavllo20193d,liu2020attention,zhang2022mixste}. We set the number of blocks of MixSTE to 4 to reduce the consumption of GPU memory.

\subsection{Experimental Results}
Table~\ref{tab:res} shows the performance results of different actions. Following previous work~\cite{pavllo20193d,liu2020attention, zhang2022mixste}, in the top group, we take the ground truth (GT) 2D pose as input to predict the 3D pose. In the bottom group, we use the HRNet~\cite{sun2019deep} as 2D pose detector to obtain 2D joints for benchmark evaluation. The improvements on the various baseline models, 2D pose types and protocols demonstrate the superior effectiveness and generality of our action-aware design.
In addition, the Tail D-MPJPE presents greater improvement than D-MPJPE for all baseline models, which means that our model can brings more benefits for poses of hard actions. The results on HumanEva-I are shown in Table~\ref{tab:humaneva}, which further verify the generalization ability of our method.

\begin{figure}[!t]
\centering
\includegraphics[scale=0.22]{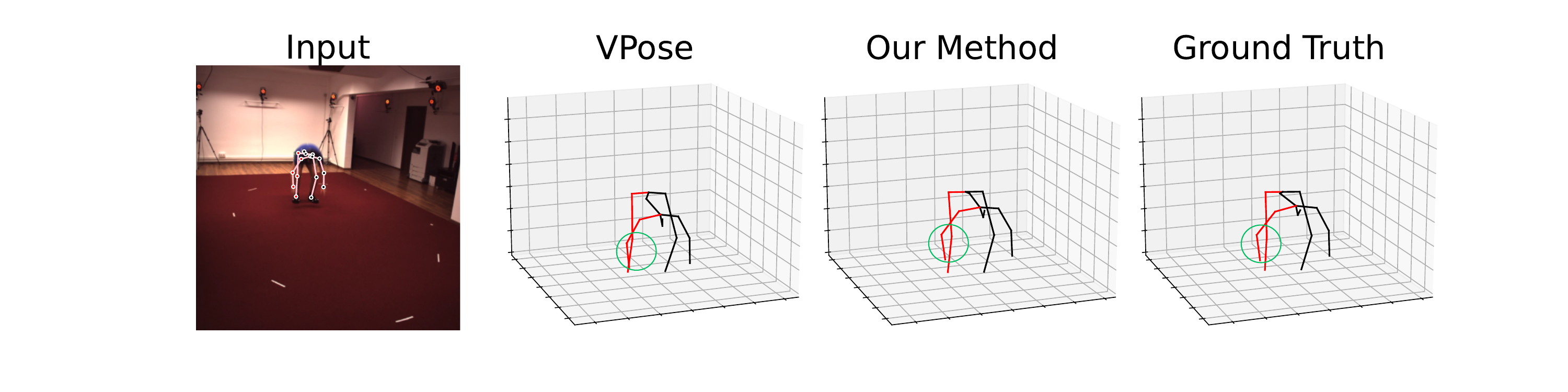}
\includegraphics[scale=0.22]{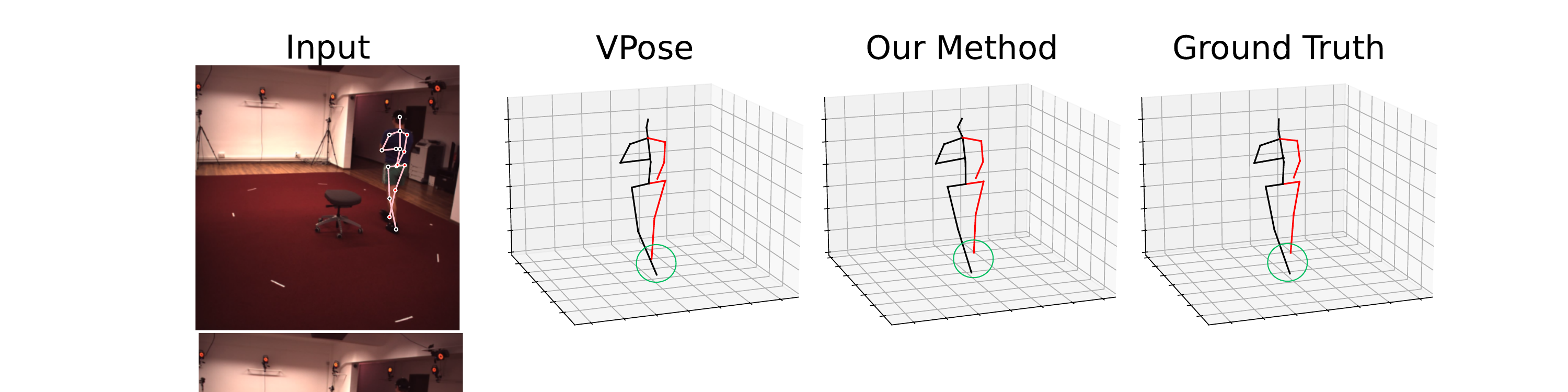}
\caption{Qualitative results on Human3.6M.}\label{fig:vis}
\end{figure}

Furthermore, Fig.~\ref{fig:res}(b) intuitively shows the improvement of D-MPJPE in different methods. The results demonstrate that our method can significantly reduce the prediction error on depth axis, especially for hard actions with higher depth ambiguity.
Fig.~\ref{fig:vis} shows the qualitative results on some hard actions in Human3.6M. Compared with baseline, our method can alleviate the depth ambiguity caused by self-occlusion.

\subsection{Ablation Studies}
Ablation studies are performed to further validate the design of each component, where APP is not added in the ablation studies on ATP. 
We take the 2D ground truth on Human3.6M as the input sequence (with an input length $F$ of 243) and VPose as the baseline. 

\begin{table}[!t]
\begin{center}
\renewcommand{\baselinestretch}{1.0}
\renewcommand{\arraystretch}{1.0}
\setlength\tabcolsep{1.2pt}
\caption{ Quantitative evaluation results  on
HumanEva-I over Walk and Jog by subject.}
\label{tab:humaneva}
\begin{tabular}{l|ccc|ccc|cc}
\hline
Method&  \multicolumn{3}{c|}{Walk} &\multicolumn{3}{c|}{Jog} &P1&  P2\\
\hline
\hline
VPose& 20.5& 15.9 &30.5 &36.1& 23.0 &25.7& \multicolumn{1}{l}{25.3}& \multicolumn{1}{l}{19.0}\\
+APM& 18.8& 16.2& 30.3 &33.1 &21.8& 24.7& 24.1\textcolor{red}{(-1.2)}  &16.6\textcolor{red}{(-2.4)}	\\
\hline
A3DHP& 17.5& 13.1& 26.4& 19.5& 17.9&  21.5& \multicolumn{1}{l}{19.3}& \multicolumn{1}{l}{12.8}\\
+APM& 16.1& 12.2& 25.4& 19.2& 16.6& 20.2& 18.3\textcolor{red}{(-1.0)}& 11.5\textcolor{red}{(-1.3)} 	\\
\hline
MixSTE&18.7& 18.0 &26.4 &27.8& 18.0 &20.0 & \multicolumn{1}{l}{21.5} & \multicolumn{1}{l}{18.1}	\\
+APM& 17.4& 17.3& 25.2 &26.6 &17.2& 18.2& 20.3\textcolor{red}{(-1.2)} &16.4\textcolor{red}{(-1.7)}	\\
\hline
\end{tabular}
\end{center}
\end{table}


\begin{figure}[!t]
\renewcommand{\baselinestretch}{1.0}
\centering
\subfloat[]{\includegraphics[scale=0.5]{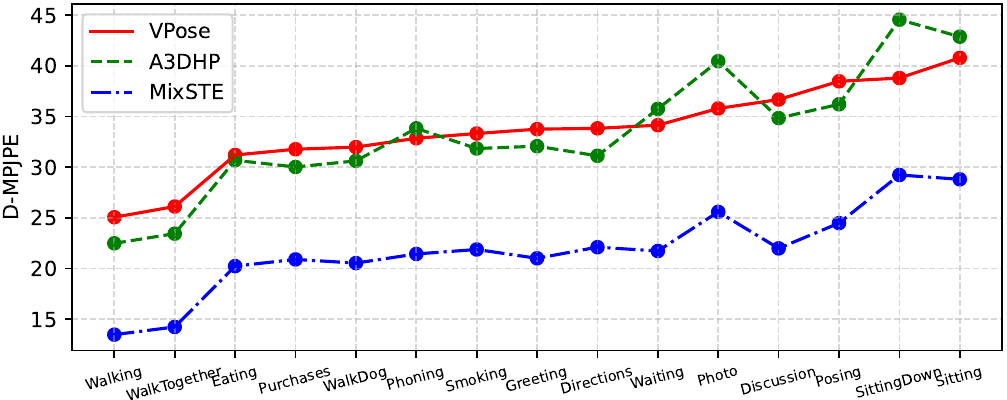}\label{fig:res_1}}\\
\vspace{-6pt}
\subfloat[]{\includegraphics[scale=0.5]{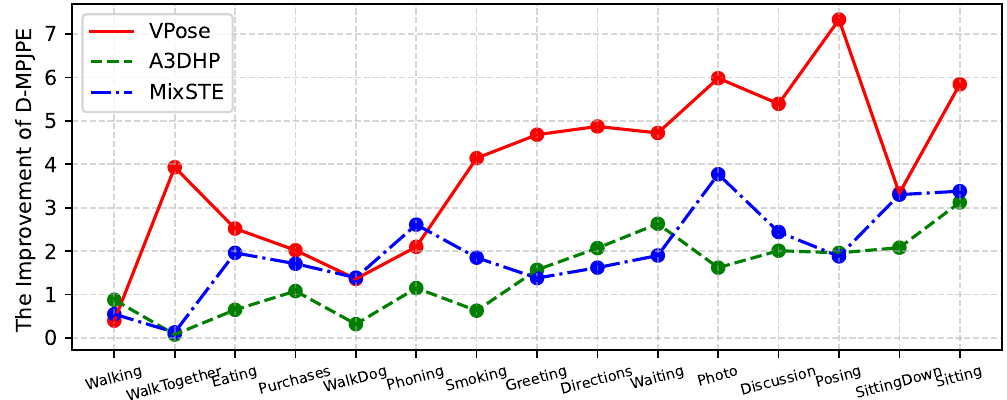}\label{fig:res_2}}
\caption{(a) D-MPJPE distribution of different actions on three baseline models. (b) Analysis of APM on hard actions. Our proposed Action Prompt Module mainly benefit hard actions with higher predicting errors.}\label{fig:res}
\end{figure}

\noindent\textbf{Each component in APM.} We first evaluate the effect of each component of our Action Prompt Module, as shown in Table~\ref{tab:component}. Firstly, we introduce the action recognition task as the sub-task using a projector the same as the Action Projector, termed as \emph{``Action Label"}. There is only 0.2mm and 1.5mm gain under two protocols, which means that multi-task learning strategy benefits 3D HPE slightly.
By adding ATP, VPose achieves 4.2\% and 10.1\% improvement under two protocols respectively, proving that the action-related language knowledge is effectively facilitated for the feature extraction of pose sequence. In addition, by adding APP and replacing ATP with the simple action projector to infer the action label,  VPose obtains 3.2\% MPJPE improvement and 6.5\% D-MPJPE improvement, which means that the mined position-aware pose pattern of each action can refine the estimated pose. Finally, by combing ATP with APP,  VPose achieves the best result boosted 5.6\%.

\noindent\textbf{Each component in ATP.} 
Since the ATP module does not depend on APP, we verified the effect of each component of ATP by removing APP for convenience.
Firstly, we adopt the simplest design for Action Projector by using global average pooling, which can benefit 3D HPE a lot already. However, we argue that amounts of temporal information is lost in this way. We introduce TCN blocks to project the pose features, which further brings 0.5mm improvement in MPJPE. Finally, further adding Pose-to-Text Prompt to refine the text prompt embeddings achieves the best result.

\noindent\textbf{K-action text prompt.} As shown in Table~\ref{tab:k-action}, we remove the text encoder, and directly set the K-action text prompt embeddings $T$ as learnable parameters.  Compared with directly using action label for multi-task learning, learnable K-action embeddings bring a better performance, while our ATP can boost more.
The results show that extra action-related information counts for 3D HPE. 
And with text encoder, we can endow more abundant information in the language domain into the pose sequence. 

\noindent\textbf{Length of input sequence.} As shown in Table~\ref{tab:input length}, we explore the effect of input sequence length. The result shows that as the input sequence length increases, the benefit from ATP gets greater. ATP achieves 2.4mm improvement in MPJPE when inputting 243 frames. It shows that long length input sequence contains long-range temporal information, which can precisely reflect the characteristics of action. In contrast, when the input length is small, the short-time frames are just a small part of the action, which contain ambiguous action-related information. Therefore, it makes sense that ATP is more suitable for long input sequence.

\begin{table}[!t]
\renewcommand{\baselinestretch}{1.0}
\renewcommand{\arraystretch}{1.0}
\centering
\caption{Ablation study on each component of Action Prompt Module. Acc denotes the  accuracy of the predicted action labels.}\label{tab:component}
\begin{tabular}{lccc|ccc}
\hline
&Action Label& ATP& APP &P1&	P2 & Acc.\\
\hline
\hline
Baseline &  & & &37.8&	33.6 & -\\
~&\checkmark& & &37.6&	32.1 & 90.2\%\\
~&\checkmark&\checkmark & &36.2&	30.2 & 94.6\%\\
~&\checkmark &&\checkmark &36.6&	31.4 &  91.5\% \\ \hline
Ours&\checkmark&\checkmark &\checkmark   &\textbf{35.7} & 	\textbf{29.7}& \textbf{96.2\%}\\
\hline
\end{tabular}
\renewcommand{\baselinestretch}{1.0}
\renewcommand{\arraystretch}{1.0}
\centering
\caption{Ablation study on each component of ATP. GP denotes global average pooling.} 
\label{tab:ATP}
\begin{tabular}{lccc|c|c}
 \hline
 &\multicolumn{2}{c}{Action Projector}&\multirow{2}*{P2T} &\multirow{2}*{P1}&	\multirow{2}*{P2}\\
 \cline{2-3}
& GP&TCN&&\\
  \hline
  \hline
Baseline& & & &37.8& 33.6 \\
&\checkmark& &  &36.9& 31.7	\\
& & \checkmark&   &36.4& 30.9	\\
&\checkmark& &\checkmark  &36.7& 31.3	\\
  \hline
Ours w/o APP& &\checkmark&\checkmark  &\textbf{36.2} &\textbf{30.2}	\\
\hline
\end{tabular}
\renewcommand{\baselinestretch}{1.0}
\renewcommand{\arraystretch}{1.0}
\centering
\caption{Ablation study on K-action text prompt. APP is added into the model.}\label{tab:k-action}
\begin{tabular}{l|cc}
\hline
Method &P1&	P2\\
\hline
\hline
Multi-task & 36.6&	31.4\\
K-action text prompt &36.3&	30.9\\
ATP (ours)&\textbf{35.7}&	\textbf{29.7}\\
\hline
\end{tabular}
\end{table}

\noindent  \textbf{Position of Action Projector.} Table~\ref{tab:proj} shows the effect of the position of Action Projector. We connect the Action Projector with different layers of  VPose. We find that as the connected layer deepens, the model performance gets worse. From the 1st layer to the 5th layer, MPJPE  increases 13.3\%. When the connected layer is 4 and 5, the extracted action information misleads the pose estimation. It can be inferred that the shallow layer features contain more general information while the deeper reflect more specific information 
related with target 3D pose, which deviates from the action recognition task.

\noindent\textbf{APP with GT action label and action prediction accuracy.} Since the action label inferred by our ATP could produce a certain error rate, as shown in Table \ref{tab:gt}, using the GT action label to select the pose prompt for adjustment can yield a slight performance gain.
We also list the accuracy of the action labels prediction for different variants of our method in Table \ref{tab:component}. We can infer that MPJPE is positively correlated with the accuracy, and our method can also boost the accuracy compared with simply introducing the action recognition task as the sub-task.

\noindent\textbf{Parameters in APP.} Table~\ref{tab:app} evaluates the impact of parameters in APP on the performance and complexity of our model. We find that using one transformer block obtains the best result and stacking more blocks does not yield gains. The result shows that enlarging the embedding dimension from 128 to 256 can boost the model performance, but cannot bring further benefits when using dimension 1024. In addition, we find that 81 pose prompts can yield the best result.

\begin{table}[!t]
\renewcommand{\baselinestretch}{1.0}
\renewcommand{\arraystretch}{1.0}
\setlength{\tabcolsep}{9pt}
\setlength{\abovecaptionskip}{0pt}
\centering
\caption{Ablation study on input sequence length.}\label{tab:input length}
\begin{tabular}{l|ccc}
\hline
Frames&  P1& P2 &  P3\\
\hline
\hline
9& \multicolumn{1}{l}{39.8}& \multicolumn{1}{l}{35.2}& \multicolumn{1}{l}{43.2}\\
9+ATP &40.2\textcolor{green}{(+0.4)}&	34.9\textcolor{red}{(-0.3)}& 42.2\textcolor{red}{(-1.0)}\\
\hline
27& \multicolumn{1}{l}{39.1}& \multicolumn{1}{l}{34.9}& \multicolumn{1}{l}{41.2}\\
27+ATP& 38.3\textcolor{red}{(-0.8)} & 32.9\textcolor{red}{(-2.0)}&38.0\textcolor{red}{(-3.2)}\\
\hline
81 &\multicolumn{1}{l}{38.1}& \multicolumn{1}{l}{33.8}& \multicolumn{1}{l}{40.4}	\\
81+ATP& 37.0\textcolor{red}{(-1.1)}&	31.5\textcolor{red}{(-2.3)}&35.5\textcolor{red}{(-4.9)}\\
  \hline
243 & \multicolumn{1}{l}{37.8}& \multicolumn{1}{l}{33.6}& \multicolumn{1}{l}{39.3}\\
243+ATP&36.2\textcolor{red}{(-1.6)}&30.2\textcolor{red}{(-3.4)}&34.1\textcolor{red}{(-5.2)}\\
\hline
\end{tabular}
\renewcommand{\baselinestretch}{1.0}
\renewcommand{\arraystretch}{1.0}
\setlength{\tabcolsep}{10pt}
\setlength{\abovecaptionskip}{0pt}
\centering
\caption{Ablation study on the position of Action Projector.}\label{tab:proj}
\begin{tabular}{l|ccc}
\hline
Layer& P1 &	P2 & P3 \\
\hline
\hline
5&	41.0\textcolor{green}{(+3.2)}&	33.9\textcolor{green}{(+0.3)}&40.8\textcolor{green}{(+1.5)}\\
4&	38.6\textcolor{green}{(+0.8)}&	32.5\textcolor{red}{(-1.1)}&39.0\textcolor{red}{(-0.3)}\\
3&	37.3\textcolor{red}{(-0.5)}&	31.2\textcolor{red}{(-2.4)}&37.5\textcolor{red}{(-1.8)}\\
2&	36.8\textcolor{red}{(-1.0)}&	30.7\textcolor{red}{(-2.9)}&37.0\textcolor{red}{(-2.3)}\\
1&	36.2\textcolor{red}{(-1.6)}&	30.2\textcolor{red}{(-3.4)}&34.1\textcolor{red}{(-5.2)}\\
\hline
\end{tabular}


\renewcommand{\baselinestretch}{1.0}
\renewcommand{\arraystretch}{1.0}
\setlength{\tabcolsep}{12pt}
\centering
\caption{Ablation study on APP with  GT action label. } 
\label{tab:gt}
\begin{tabular}{l|cc}
\hline
Method &P1&	P2\\
\hline
\hline
APP w. GT & \textbf{36.4}&	\textbf{31.0}\\
APP w/o. GT &36.6&	31.4\\
\hline
\end{tabular}
\renewcommand{\baselinestretch}{1.0}
\renewcommand{\arraystretch}{1.0}
\centering
\caption{Ablation study on different parameters of APP. $C$ is the embedding dimension. $D$  and $L$ denote the number of  APP layers and pose prompts, respectively.}\label{tab:app}
\begin{tabular}{ccc|c|c|c}
\hline
D& C& L& P1 &P2 & Params(M) \\
\hline
\hline
1& 256 &81 &	\textbf{35.7}& 29.7&	22.87\\
4& 256 &81 &	35.8&\textbf{29.3}&	24.45\\
  \hline
1& 128 &81 &	36.0&29.4&	22.33\\
1& 1024 &81 &	35.8&29.5&	31.61\\
  \hline
1& 256 &27 &	36.2&29.9&	22.76\\
1& 256 &243 &	36.3&30.1&	23.21\\
\hline
\end{tabular}
\end{table}

\section{Conclusion}
In this paper, we proposed a plug-and-play module, named Action Prompt Module (APM), to mine action clues for 3D HPE. We first present a novel Action-related Text Prompt module (ATP) that adapts the rich action-related language information in action label to the pose sequence. 
Secondly, to mine the position-aware pose pattern of each action, we introduce
Action-specific Pose Prompt (APP), which refines pose feature by exploiting the correlation between learnable patterns with input pose sequence.
APM can be applicable to the most video-based 2D-to-3D HPE methods, and extensive results on Human3.6M and HumanEva-I reveal the benefits of our design for  3D pose encoders. 

\section{Acknowledgement}
This work was supported in part by the National Natural Science Foundation of China under Grants 61932022, 61931023, 61971285, 61831018, and 61972256. 

\bibliographystyle{IEEEtran}
{\small \bibliography{refer}}


\end{document}